\documentclass{article}

 \usepackage[preprint]{neurips_2025}

\usepackage[utf8]{inputenc} 
\usepackage[T1]{fontenc}    
\usepackage{hyperref}       
\usepackage{url}            
\usepackage{booktabs}       
\usepackage{amsfonts}       
\usepackage{nicefrac}       
\usepackage{microtype}      

\usepackage{tikz}
\usetikzlibrary{arrows.meta, positioning, fit, shapes.geometric, calc}

\tikzset{
  tiny/.style={font=\tiny},
  small/.style={font=\scriptsize},
  box/.style={rounded corners=2pt, draw, align=center, inner xsep=1pt,
  inner ysep=3pt, outer sep=1pt},
  data/.style={box, fill=gray!15, draw=gray!60},
  synth/.style={box, fill=cyan!15, draw=cyan!60},
  final/.style={box, fill=green!20, draw=green!70, thick},
  teacher/.style={box, fill=purple!15, draw=purple!60},
  logits/.style={box, fill=red!10, draw=red!40},
  arrow/.style={-{Latex[scale=0.8]}, line width=0.5pt},
  varrow/.style={-{Latex[scale=0.8]}, dashed, line width=0.4pt, gray},
  nodew/.style={text width=32mm, minimum height=16mm},
  panel/.style={draw=black!30, rounded corners=3pt, inner sep=4pt}
}

\usepackage{xcolor}         

\makeatletter
\renewcommand{\@listi}{\leftmargin\leftmargini
  \topsep 4pt plus 1pt minus 2pt     
  \parsep 2pt plus 1pt minus 1pt     
  \itemsep 2pt plus 1pt minus 1pt}   
\makeatother

\title{How to Train Private Clinical Language Models: A Comparative Study of Privacy-Preserving Pipelines for ICD-9 Coding}
\workshoptitle{Privacy-Preserving Machine Learning Workshop}

\author{%
  Mathieu Dufour \\
  Department of Mathematics \\
  Imperial College London \\
  \texttt{mathieu.dufour23@imperial.ac.uk} \\
  \And
  Andrew Duncan \\
  Department of Mathematics \\
  Imperial College London \\
  \texttt{a.duncan@imperial.ac.uk} \\
}

\begin{document}

\maketitle

\begin{abstract}
Large language models trained on clinical text risk exposing sensitive patient information, yet differential privacy (DP) methods often severely degrade the diagnostic accuracy needed for deployment. Despite rapid progress in DP optimisation and text generation, it remains unclear which privacy-preserving strategy actually works best for clinical language tasks. We present the first systematic head-to-head comparison of four training pipelines for automated diagnostic coding from hospital discharge summaries. All pipelines use identical 1B-parameter models and matched privacy budgets to predict ICD-9 codes. At moderate and relaxed privacy budgets ($\varepsilon \in \{4, 6\}$), knowledge distillation from DP-trained teachers outperforms both direct DP-SGD and DP-synthetic data training, recovering up to 63\% of the non-private performance whilst maintaining strong empirical privacy (membership-inference AUC $\approx$ 0.5). These findings expose large differences in the privacy-utility trade-off across architectures and identify knowledge distillation as the most practical route to privacy-preserving clinical NLP.
\end{abstract}

\section{Introduction}

Large language models trained on clinical text can inadvertently memorise and reveal sensitive patient details \citep{carlini2021extractingtrainingdatalarge}, raising serious concerns about their deployment in healthcare. Differential privacy (DP) offers formal guarantees that individual patient records cannot be reliably inferred from model outputs \citep{dwork2014algorithmic}, but applying DP training to large models often comes at a steep price: accuracy drops of 40\% or more have been reported for clinical NLP tasks \citep{dadsetan2024largelanguagemodelsprivacy}. This tension between diagnostic utility and patient confidentiality has become a central challenge for trustworthy clinical AI.

A variety of strategies have been proposed to mitigate these trade-offs. Parameter-efficient tuning methods such as Low-Rank Adaptation (LoRA) restrict memorisation through low-rank updates and may offer partial implicit privacy \citep{malekmohammadi2025lowrankadaptationsecretlyimitates}. Generative approaches train large DP-protected models to produce synthetic clinical notes that can be used freely for downstream tasks \citep{yue2023synthetictextgenerationdifferential}. More recently, differentially private knowledge distillation allows a private ``teacher'' to transfer its knowledge to a smaller ``student'' via synthetic data and soft labels, preserving the teacher's privacy guarantee through post-processing \citep{flemings2024differentiallyprivateknowledgedistillation}.

Despite this progress, no prior work has compared these methods under identical conditions. Existing studies vary in model size, dataset, or privacy accounting, making it unclear which approach provides the best balance of accuracy, efficiency, and protection. For practitioners deciding how to deploy language models in hospitals, this lack of systematic evidence leaves a crucial gap.

This study asks a single question: ``Given a fixed model capacity, which privacy mechanism best balances diagnostic accuracy and formal privacy guarantees in clinical NLP?''

We answer this through a controlled, head-to-head comparison of four privacy-preserving training pipelines for automated ICD-9 diagnostic coding, i.e. the task of mapping hospital discharge summaries to standardised disease codes, using the publicly available MIMIC-III dataset \citep{johnson2016mimic}. All final classifiers share the same architecture and capacity (1B parameters) to isolate the impact of the privacy mechanism itself. In pipelines involving teacher-student architectures (DP-Synthetic, DP-Distil), we train larger 3B ``teacher'' models under DP-SGD, then use their synthetic outputs and/or soft labels to train the final 1B ``student'' classifier. Because the student only accesses post-processed outputs of the DP teachers, it inherits the same formal privacy guarantees without direct exposure to real patient records.

The evaluated pipelines are: \textbf{(1) DP-Small}—direct DP-SGD on the 1B model; \textbf{(2) DP-Synthetic}—training on synthetic notes from a 3B DP-protected generator; \textbf{(3) DP-Distil}—knowledge distillation from DP-trained 3B teachers providing synthetic data and soft labels; \textbf{(4) LoRA-No-DP}—non-private LoRA fine-tuning as a utility upper bound.

We assess each approach across privacy budgets $\varepsilon \in \{2, 4, 6\}$ with fixed $\delta = 10^{-5}$, measuring utility (micro/macro-F$_1$, AUPRC) and empirical privacy via membership-inference attacks \citep{carlini2021mia}. By holding model capacity constant, our study isolates how different training pipelines reshape the privacy-utility frontier in clinical NLP. The results reveal that architectural choices, not merely the privacy budget, substantially shape this trade-off, highlighting knowledge distillation as an effective pathway for developing differentially private language models that preserve clinical usefulness.

\section{Methods}

\subsection{Dataset}

We evaluate all methods on the MIMIC-III v1.4 critical-care database \citep{johnson2016mimic}, a collection of de-identified electronic health records from intensive care unit admissions. We use discharge summaries—clinical narratives summarising each hospital stay—associated with the 50 most frequent ICD-9 diagnostic codes. ICD-9 (International Classification of Diseases, 9th Revision) is a standardised medical coding system used for billing and epidemiology. The final dataset contains 44,778 training notes, 5,624 validation notes, and 5,586 test notes, with splits defined at the admission level to prevent patient overlap. Notes are tokenised using the LLaMA-3.2 tokeniser and truncated to 512 tokens. The prediction task is multi-label ICD-9 classification.

\subsection{Model Architecture}

All final classifiers share an identical 1B-parameter LLaMA-3.2 backbone \citep{touvron2023llama, grattafiori2024llama3}, fine-tuned via LoRA \citep{hu2021loralowrankadaptationlarge} on the query (Q) and value (V) projections (rank $r=4$, $\alpha=16$). The DP ``teacher'' and ``generator'' models use the 3B-parameter variant with LoRA rank $r=8$ and $\alpha=32$. These settings yield approximately 0.53M trainable parameters for 1B models and 2.3--2.4M for 3B models, keeping adapter capacity roughly proportional to 
backbone size and ensuring a controlled comparison across training pipelines. This design follows empirical observations that larger models tend to retain higher utility under DP-SGD at fixed privacy budgets \citep{kamath2022dplm}.

We standardise on 1B students rather than deploying 3B teachers directly for two reasons: (1) to ensure fair comparison across pipelines at identical capacity, and (2) to enable resource-efficient deployment. The 1B models offer 2.7× faster inference at 13ms versus 36ms per sample and require only 4.4GB VRAM compared to 8.5GB for the teachers (Table~\ref{tab:inference-appendix}), making them practical for clinical settings with hardware constraints.

\subsection{Differential privacy mechanisms}

Differentially private training follows DP-SGD \citep{abadi2016deep} with per-example gradient clipping and Gaussian noise calibrated via R\'enyi DP accounting in Opacus \citep{yousefpour2021opacus}. Unless stated otherwise, $\delta = 10^{-5}$ and $\varepsilon \in \{2, 4, 6\}$, corresponding to strong, moderate, and relaxed privacy regimes commonly reported in medical NLP \citep{dadsetan2024largelanguagemodelsprivacy}. Gradient clipping thresholds are $C=1.0$ for 1B models and $C=0.7$ for 3B models. In DP-Distil, each teacher model consumes $\varepsilon/2$ to preserve overall $(\varepsilon, \delta)$ under sequential composition \citep{dwork2014algorithmic}.

\subsection{Privacy-preserving training pipelines}

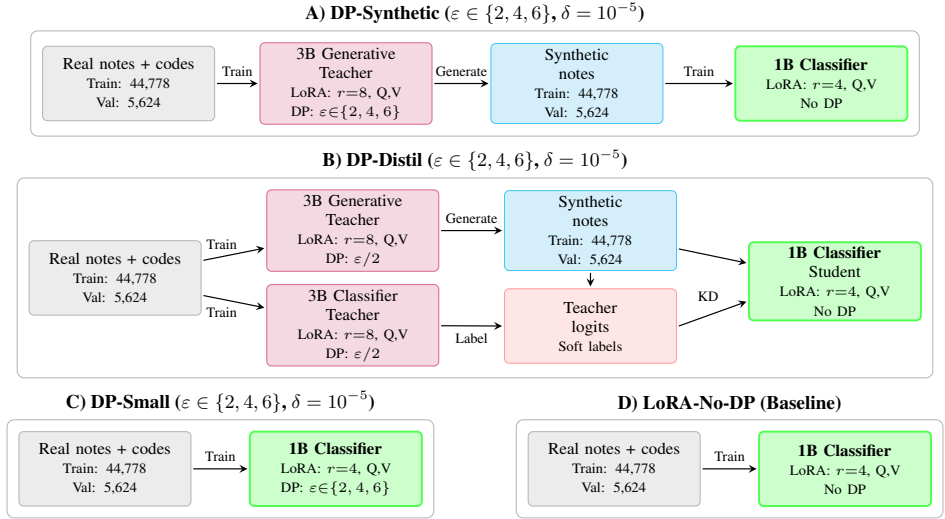
\begin{figure}[t]
\centering
\footnotesize
\tikzset{
  nodew/.style={text width=24mm, minimum height=11mm},
  box/.style={rounded corners=2pt, draw, align=center, inner sep=2pt, outer sep=1pt},
  data/.style={box, fill=gray!15, draw=gray!60},
  synth/.style={box, fill=cyan!15, draw=cyan!60},
  final/.style={box, fill=green!20, draw=green!70, thick},
  teacher/.style={box, fill=purple!15, draw=purple!60},
  logits/.style={box, fill=red!10, draw=red!40},
  arrow/.style={->, >=stealth, line width=0.5pt},
  panel/.style={draw=black!30, rounded corners=3pt, inner sep=4pt},
  small/.style={font=\scriptsize}
}

\scalebox{0.90}{
\begin{tikzpicture}[node distance=3mm and 6mm, small]
\node[data, nodew] (realA) {Real notes + codes\\{\tiny Train: 44,778}\\{\tiny Val: 5,624}};
\node[teacher, nodew, right=of realA] (teachA) {3B Generative\\Teacher\\{\tiny LoRA: $r{=}8$, Q,V}\\{\tiny DP: $\varepsilon{\in}\{2,4,6\}$}};
\node[synth, nodew, right=8mm of teachA] (synthA) {Synthetic\\notes\\{\tiny Train: 44,778}\\{\tiny Val: 5,624}}; 
\node[final, nodew, right=10mm of synthA] (finalA) {\textbf{1B Classifier}\\{\tiny LoRA: $r{=}4$, Q,V}\\{\tiny No DP}};
\draw[arrow] (realA) -- node[above, font=\tiny] {Train} (teachA);
\draw[arrow] (teachA) -- node[above, font=\tiny] {Generate} (synthA);
\draw[arrow] (synthA) -- node[above, font=\tiny] {Train} (finalA);
\node[panel, fit=(realA)(teachA)(synthA)(finalA),
      label={[font=\footnotesize\bfseries, scale=0.9]above:A) DP-Synthetic ($\varepsilon \in \{2,4,6\}$, $\delta=10^{-5}$)}] {};
\end{tikzpicture}
}

\vspace{0.5mm}

\scalebox{0.90}{
\begin{tikzpicture}[node distance=3mm and 6mm, small]
\node[data, nodew] (realB) {Real notes + codes\\{\tiny Train: 44,778}\\{\tiny Val: 5,624}};
\node[teacher, nodew, right=of realB, xshift=3mm, yshift=7mm] (teachGenB) {3B Generative\\Teacher\\{\tiny LoRA: $r{=}8$, Q,V}\\{\tiny DP: $\varepsilon/2$}};
\node[teacher, nodew, right=of realB, xshift=3mm, yshift=-7mm] (teachClassB) {3B Classifier\\Teacher\\{\tiny LoRA: $r{=}8$, Q,V}\\{\tiny DP: $\varepsilon/2$}};
\node[synth, nodew, right=of teachGenB, xshift=3mm] (synthB) {Synthetic\\notes\\{\tiny Train: 44,778}\\{\tiny Val: 5,624}};
\node[logits, nodew, below=2mm of synthB] (logitsB) {Teacher\\logits\\{\tiny Soft labels}};
\node[final, nodew, right=10mm of synthB, yshift=-7.5mm] (finalB) {\textbf{1B Classifier}\\Student\\{\tiny LoRA: $r{=}4$, Q,V}\\{\tiny No DP}};

\draw[arrow] (realB) -- node[above, font=\tiny, pos=0.3] {Train} (teachGenB);
\draw[arrow] (realB) -- node[below, font=\tiny, pos=0.3] {Train} (teachClassB);
\draw[arrow] (teachGenB) -- node[above, font=\tiny] {Generate} (synthB);
\draw[arrow] (teachClassB) -- node[below, font=\tiny] {Label} ([yshift=0]logitsB.west);
\draw[arrow] (synthB) -- (logitsB);
\draw[arrow] (logitsB.east) -- node[above left, font=\tiny, pos=0.7] {KD} ([yshift=-3mm]finalB.west);
\draw[arrow] (synthB) -- (finalB);

\node[panel, fit=(realB)(teachGenB)(teachClassB)(synthB)(logitsB)(finalB),
      label={[font=\footnotesize\bfseries, scale=0.9]above:B) DP-Distil ($\varepsilon \in \{2,4,6\}$, $\delta=10^{-5}$)}] {};
\end{tikzpicture}
}

\vspace{0.5mm}

\scalebox{0.90}{
\begin{tikzpicture}[node distance=3mm and 8mm, small]
\node[data, nodew] (realC) {Real notes + codes\\{\tiny Train: 44,778}\\{\tiny Val: 5,624}};
\node[final, nodew, right=of realC] (finalC) {\textbf{1B Classifier}\\{\tiny LoRA: $r{=}4$, Q,V}\\{\tiny DP: $\varepsilon{\in}\{2,4,6\}$}};
\draw[arrow] (realC) -- node[above, font=\tiny] {Train} (finalC);
\node[panel, fit=(realC)(finalC),
      label={[font=\footnotesize\bfseries, scale=0.9]above:C) DP-Small ($\varepsilon \in \{2,4,6\}$, $\delta=10^{-5}$)}] (panelC) {};

\node[data, nodew, right=15mm of finalC] (realD) {Real notes + codes\\{\tiny Train: 44,778}\\{\tiny Val: 5,624}};
\node[final, nodew, right=of realD] (finalD) {\textbf{1B Classifier}\\{\tiny LoRA: $r{=}4$, Q,V}\\{\tiny No DP}};
\draw[arrow] (realD) -- node[above, font=\tiny] {Train} (finalD);
\node[panel, fit=(realD)(finalD),
      label={[font=\footnotesize\bfseries, scale=0.9]above:D) LoRA-No-DP (Baseline)}] {};
\end{tikzpicture}
}

\caption{Four training pipelines producing identical 1B classifiers (green). 
\textbf{(A)} DP-trained 3B generator creates synthetic data. 
\textbf{(B)} DP-trained 3B teachers (each $\varepsilon/2$) provide synthetic data and soft labels for knowledge distillation. 
\textbf{(C)} Direct DP-SGD on 1B model. 
\textbf{(D)} Non-private LoRA baseline. 
All DP pipelines use $\varepsilon \in \{2, 4, 6\}$ with $\delta = 10^{-5}$.}
\label{fig:pipelines}
\vspace{-1mm}
\end{figure}

Figure~\ref{fig:pipelines} illustrates the four pipelines, all producing identical 1B classifiers for fair comparison.

\textbf{DP-Synthetic}: A 3B generative model is fine-tuned under DP-SGD for conditional generation with control codes \citep{yue2023synthetictextgenerationdifferential}, producing 44,778 synthetic notes (nucleus sampling: $p=0.9$, $T=0.8$) to train a 1B classifier inheriting the generator's $(\varepsilon, \delta)$ via post-processing.

\textbf{DP-Distil}: Two 3B teachers trained under DP-SGD (each $\varepsilon/2, \delta/2$)—generative and classification—provide synthetic data and soft labels. A 1B student trains via MSE on raw logits ($\alpha=0$), inheriting $(\varepsilon, \delta)$ through sequential composition \citep{flemings2024differentiallyprivateknowledgedistillation}.

\textbf{DP-Small}: Direct DP-SGD training of the 1B model with binary cross-entropy loss and frequency-weighted positive class weights, evaluating direct private training without architectural modifications.

\textbf{LoRA-No-DP}: LoRA fine-tuning without gradient clipping or noise, serving as an upper bound on utility and testing LoRA's implicit privacy properties \citep{malekmohammadi2025lowrankadaptationsecretlyimitates}.

\subsection{Evaluation}

\textbf{Utility.} We report Micro- and Macro-F$_1$, Micro-AUPRC, and Hamming loss, with thresholds optimised on the validation set. Micro-F$_1$ aggregates predictions across all labels, whilst Macro-F$_1$ computes per-label F$_1$ scores and averages them, giving equal weight to rare codes.

\textbf{Privacy.} Following \citet{carlini2021mia} and \citet{shokri2017membershipinferenceattacksmachine}, we train logistic-regression membership-inference attacks on five features derived from model logits: cross-entropy loss, maximum confidence, entropy, confidence margin, and L$_2$ norm. Attacks are trained on a balanced dataset of 5,586 training members versus 5,586 test non-members. AUC = 0.5 indicates random guessing.

\section{Results}

\subsection{Overall performance}

Table~\ref{tab:main-results} summarises classification accuracy across all privacy budgets $\varepsilon \in \{2, 4, 6\}$ ($\delta = 10^{-5}$). At the strictest budget ($\varepsilon = 2$), DP-Small performs best among DP methods (Micro-F$_1$ $\approx$ 0.26), reflecting the advantage of allocating the full privacy budget to a single model. As $\varepsilon$ increases, DP-Distil overtakes it, reaching 0.33 Micro-F$_1$ at $\varepsilon = 6$, an 8.7\% gain over DP-Small and a 48\% improvement over DP-Synthetic. The non-private LoRA-No-DP baseline achieves 0.52, defining the practical upper bound on utility.

\begingroup
\setlength{\tabcolsep}{4.5pt} 
\renewcommand{\arraystretch}{0.95}
\scriptsize
\begin{table}[h!]
\centering
\caption{Classification performance across pipelines and privacy budgets ($\delta = 10^{-5}$ for all DP methods). Arrows indicate direction of improvement.}
\label{tab:main-results}
\small
\begin{tabular}{llrrrr}
\toprule
\textbf{Pipeline} & \textbf{$\varepsilon$} & \textbf{Micro-F$_1$ $\uparrow$} & \textbf{Macro-F$_1$ $\uparrow$} & \textbf{Micro-AUPRC $\uparrow$} & \textbf{Ham $\downarrow$} \\
\midrule
DP-Distil & 2 & 0.217 & 0.232 & 0.260 & 0.533 \\
DP-Small & 2 & \textbf{0.258} & \textbf{0.257} & \textbf{0.285} & \textbf{0.343} \\
DP-Synthetic & 2 & 0.220 & 0.207 & 0.198 & 0.402 \\
\midrule
DP-Distil & 4 & \textbf{0.289} & \textbf{0.280} & \textbf{0.321} & \textbf{0.273} \\
DP-Small & 4 & 0.279 & 0.275 & 0.305 & 0.286 \\
DP-Synthetic & 4 & 0.228 & 0.216 & 0.212 & 0.442 \\
\midrule
DP-Distil & 6 & \textbf{0.330} & \textbf{0.303} & \textbf{0.347} & \textbf{0.216} \\
DP-Small & 6 & 0.304 & 0.298 & 0.334 & 0.247 \\
DP-Synthetic & 6 & 0.222 & 0.213 & 0.225 & 0.476 \\
\midrule
LoRA-No-DP & $\infty$ & \textbf{0.521} & \textbf{0.499} & \textbf{0.565} & \textbf{0.099} \\
\bottomrule
\end{tabular}
\end{table}
\endgroup

\begin{figure}[t]
\centering
\includegraphics[width=0.75\textwidth]{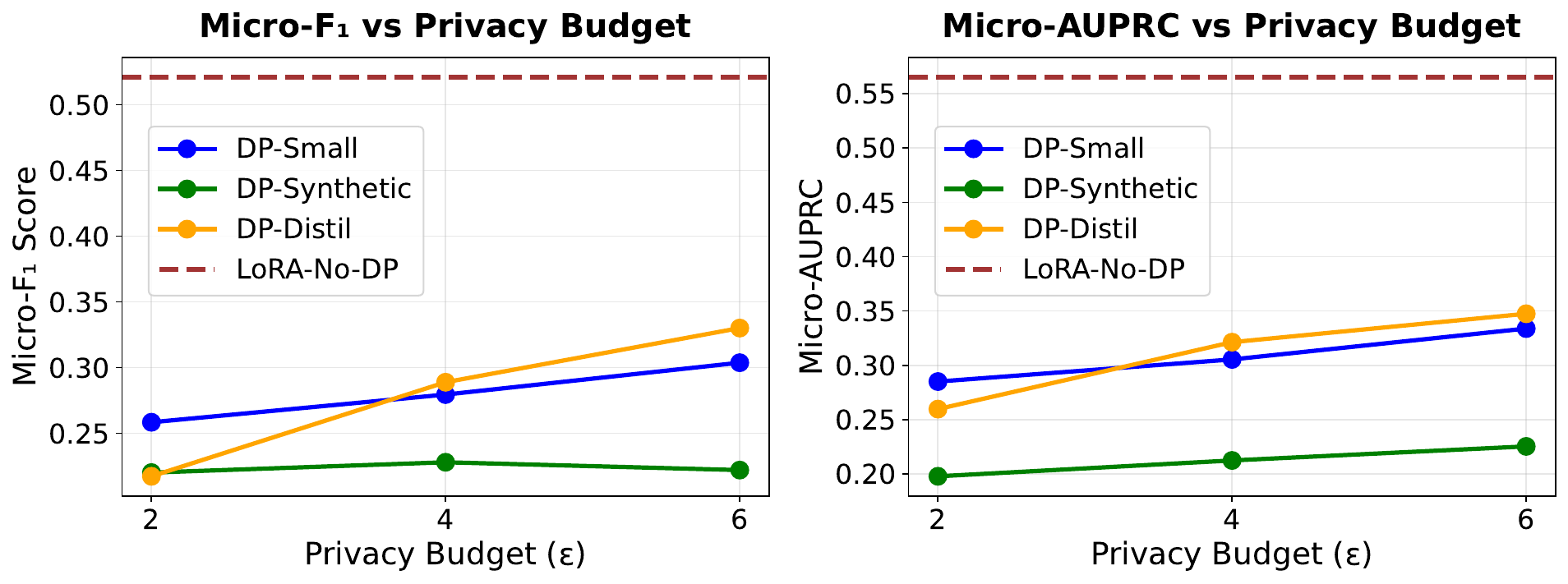}
\caption{Utility-privacy trade-off for Micro-F$_1$ (left) and Micro-AUPRC (right). DP-Distil and DP-Small improve monotonically with privacy budget, whilst DP-Synthetic plateaus regardless of $\varepsilon$. DP-Distil dominates at $\varepsilon \geq 4$; DP-Synthetic fails to scale.}
\label{fig:utility-privacy-tradeoff}
\end{figure}

Performance scales monotonically with privacy budget for DP-Small and DP-Distil but remains flat for DP-Synthetic (Figure~\ref{fig:utility-privacy-tradeoff}, $\approx$ 0.22 F$_1$ across $\varepsilon$). At $\varepsilon = 6$, DP-Distil recovers 63\% of the non-private performance whilst maintaining formal DP guarantees, demonstrating that architectural design can offset much of the utility loss induced by DP noise.

\subsection{Privacy evaluation}

Table~\ref{tab:privacy-results} reports membership-inference attack (MIA) results. All DP methods yield AUCs between 0.49 and 0.51, effectively random guessing, confirming strong empirical privacy. The LoRA-No-DP baseline, whilst partially protective (AUC $\approx$ 0.57), remains measurably vulnerable, notably to loss-based attacks (AUC $\approx$ 0.54).

\begingroup
\setlength{\tabcolsep}{4.5pt} 
\renewcommand{\arraystretch}{0.95}
\scriptsize
\begin{table}[h!]
\centering
\caption{Privacy evaluation via membership inference attacks across all pipelines and privacy budgets, showing results for ensemble attacks and individual attack features (loss, confidence, entropy, margin). Values closer to 0.5 indicate better privacy (random guessing). All DP methods achieve AUC $\approx$ 0.5; LoRA-No-DP shows vulnerability with ensemble AUC of 0.565.}
\label{tab:privacy-results}
\small
\begin{tabular}{llrrrrr}
\toprule
\textbf{Pipeline} & \textbf{$\varepsilon$} & \textbf{Ensemble} & \textbf{Loss} & \textbf{Conf.} & \textbf{Entropy} & \textbf{Margin} \\
& & \textbf{AUC} & \textbf{AUC} & \textbf{AUC} & \textbf{AUC} & \textbf{AUC} \\
\midrule
DP-Distil & 2 & 0.487 & 0.499 & 0.500 & 0.500 & 0.500 \\
DP-Distil & 4 & 0.485 & 0.504 & 0.497 & 0.502 & 0.501 \\
DP-Distil & 6 & 0.498 & 0.506 & 0.492 & 0.504 & 0.499 \\
\midrule
DP-Small & 2 & 0.512 & 0.505 & 0.492 & 0.509 & 0.498 \\
DP-Small & 4 & 0.497 & 0.507 & 0.493 & 0.507 & 0.503 \\
DP-Small & 6 & 0.499 & 0.505 & 0.494 & 0.508 & 0.493 \\
\midrule
DP-Synthetic & 2 & 0.506 & 0.500 & 0.498 & 0.503 & 0.503 \\
DP-Synthetic & 4 & 0.496 & 0.503 & 0.495 & 0.505 & 0.498 \\
DP-Synthetic & 6 & 0.504 & 0.501 & 0.502 & 0.503 & 0.507 \\
\midrule
LoRA-No-DP & $\infty$ & 0.565 & 0.536 & 0.493 & 0.507 & 0.510 \\
\bottomrule
\end{tabular}
\end{table}
\endgroup

\begin{figure}[t]
\centering
\includegraphics[width=0.9\textwidth]{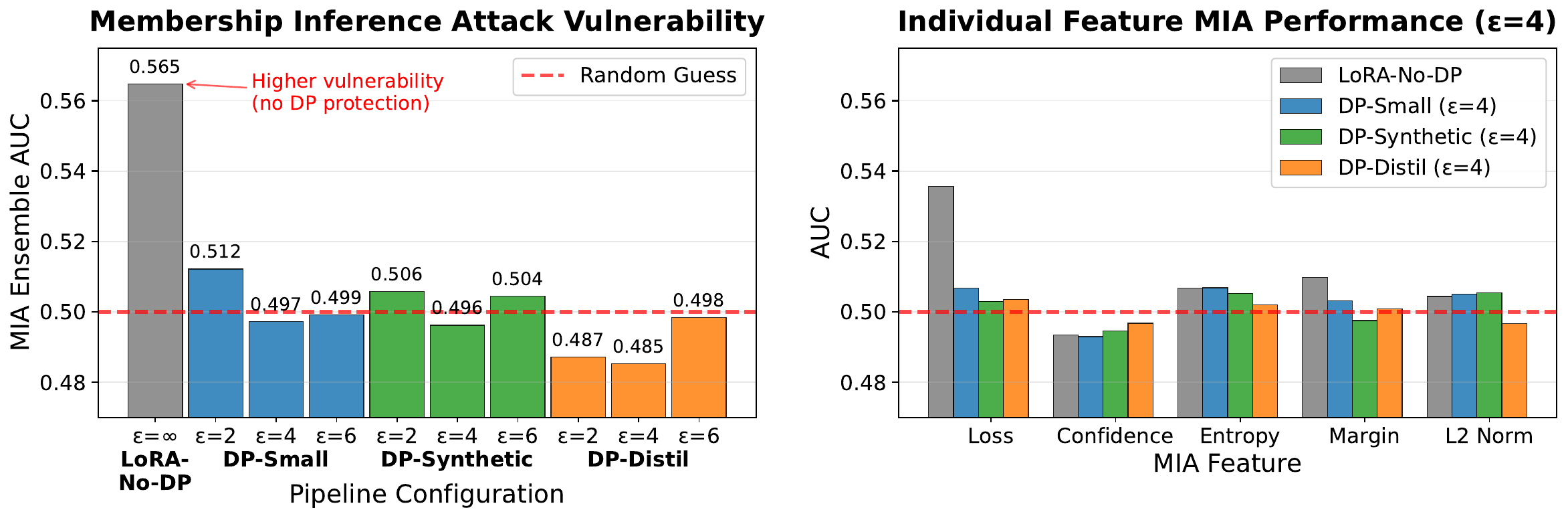}
\caption{MIA vulnerability analysis. Left: Ensemble AUC showing LoRA-No-DP's higher vulnerability versus near-random performance for DP methods. Right: Individual feature contributions at $\varepsilon=4$, demonstrating consistent protection across attack vectors for DP methods. All DP methods achieve AUC $\approx$ 0.5; LoRA-No-DP remains vulnerable.}
\label{fig:mia-vulnerability}
\end{figure}

Figure~\ref{fig:mia-vulnerability} shows ensemble MIA performance across configurations (left) and individual feature decomposition at $\varepsilon=4$ (right). For DP models, all features—loss, confidence, entropy, margin, and L$_2$ norm—cluster around 0.5 AUC, indicating that differential privacy neutralises multiple leakage channels simultaneously. Notably, empirical AUC shows no systematic correlation with $\varepsilon$, suggesting that, within this range, the presence of DP noise itself matters more than the precise choice of $\varepsilon$.

\subsection{Training efficiency and deployment considerations}

While DP-Distil achieves the best utility-privacy trade-off, this comes at a computational cost. Training times vary substantially across pipelines: DP-Distil requires 38–49 hours due to training two 3B teachers plus the 1B student, compared to 14–15 hours for DP-Synthetic, 5–10 hours for DP-Small, and just 3 hours for LoRA-No-DP (all on RTX 6000 Ada GPUs, see Table~\ref{tab:train-time-appendix} for details).

However, this training overhead is a one-time cost. All pipelines produce architecturally identical 1B classifiers that require only 13ms per sample and 4.4GB VRAM at inference (Table~\ref{tab:inference-appendix}). This uniformity means that choosing DP-Distil does not impose any deployment penalties—the same efficient 1B model runs in production regardless of how it was trained. For healthcare institutions processing thousands of daily predictions, the initial training investment in DP-Distil may be justified by its sustained 8.7\% accuracy advantage over DP-Small at $\varepsilon = 6$.

\subsection{Teacher-student distillation performance}

\begin{figure}[h!]
\centering
\includegraphics[width=0.85\textwidth]{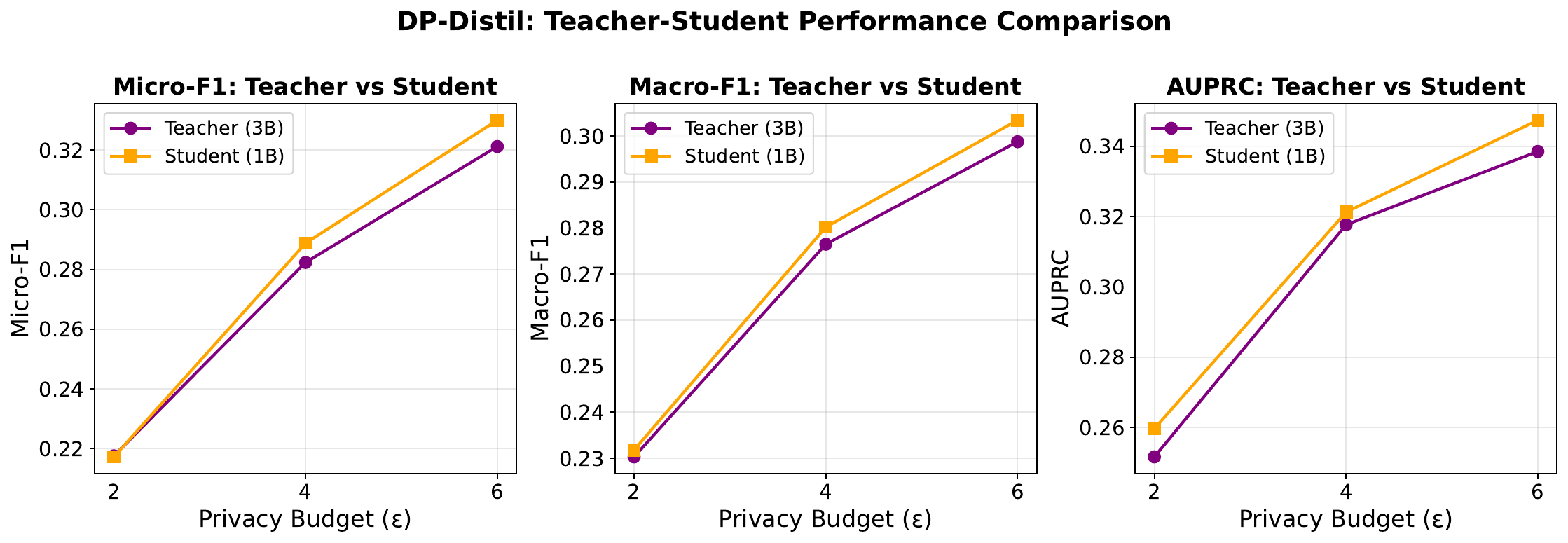}
\caption{Performance comparison between 3B-parameter DP-Distil teachers 
and their 1B-parameter distilled students across privacy budgets 
($\varepsilon \in \{2, 4, 6\}$).}
\label{fig:teacher-student}
\end{figure}

Table~\ref{tab:teacher-student-comparison} shows that 1B-parameter students consistently match or slightly outperform their 3B-parameter teachers in DP-Distil. At $\varepsilon = 6$, the student achieves Micro-F$_1$ of 0.3300 versus the teacher's 0.3212, with similar patterns for Macro-F$_1$. This result is consistent with previous distillation literature showing students can surpass teachers through regularisation effects \citep{furlanello2018bornneuralnetworks, nagarajan2024studentteacherdeviationsdistillationdoes}. The student advantage is most pronounced at higher privacy budgets where teachers have more utility to transfer, while the gap narrows at stricter settings ($\varepsilon = 2$: +0.0004 for Micro-F$_1$), as visualised in Figure~\ref{fig:teacher-student}.

\begin{table}[h!]
\centering
\caption{Performance comparison between DP-Distil teachers (3B parameters) and their distilled students (1B parameters). Positive gaps indicate teacher superiority.}
\label{tab:teacher-student-comparison}
\small
\begin{tabular}{lrrrrrr}
\toprule
\textbf{$\varepsilon$} & \textbf{Teacher} & \textbf{Student} & \textbf{Gap} & \textbf{Teacher} & \textbf{Student} & \textbf{Gap} \\
 & \multicolumn{3}{c}{\textbf{Micro-F$_1$}} & \multicolumn{3}{c}{\textbf{Macro-F$_1$}} \\
\midrule
2 & 0.2176 & 0.2172 & 0.0004 & 0.2303 & 0.2318 & -0.0015 \\
4 & 0.2824 & 0.2888 & -0.0064 & 0.2765 & 0.2802 & -0.0037 \\
6 & 0.3212 & 0.3300 & -0.0088 & 0.2988 & 0.3034 & -0.0047 \\
\bottomrule
\end{tabular}
\end{table}

\subsection{Per-label performance}

\begin{figure}[t]
\centering
\includegraphics[width=0.85\textwidth]
{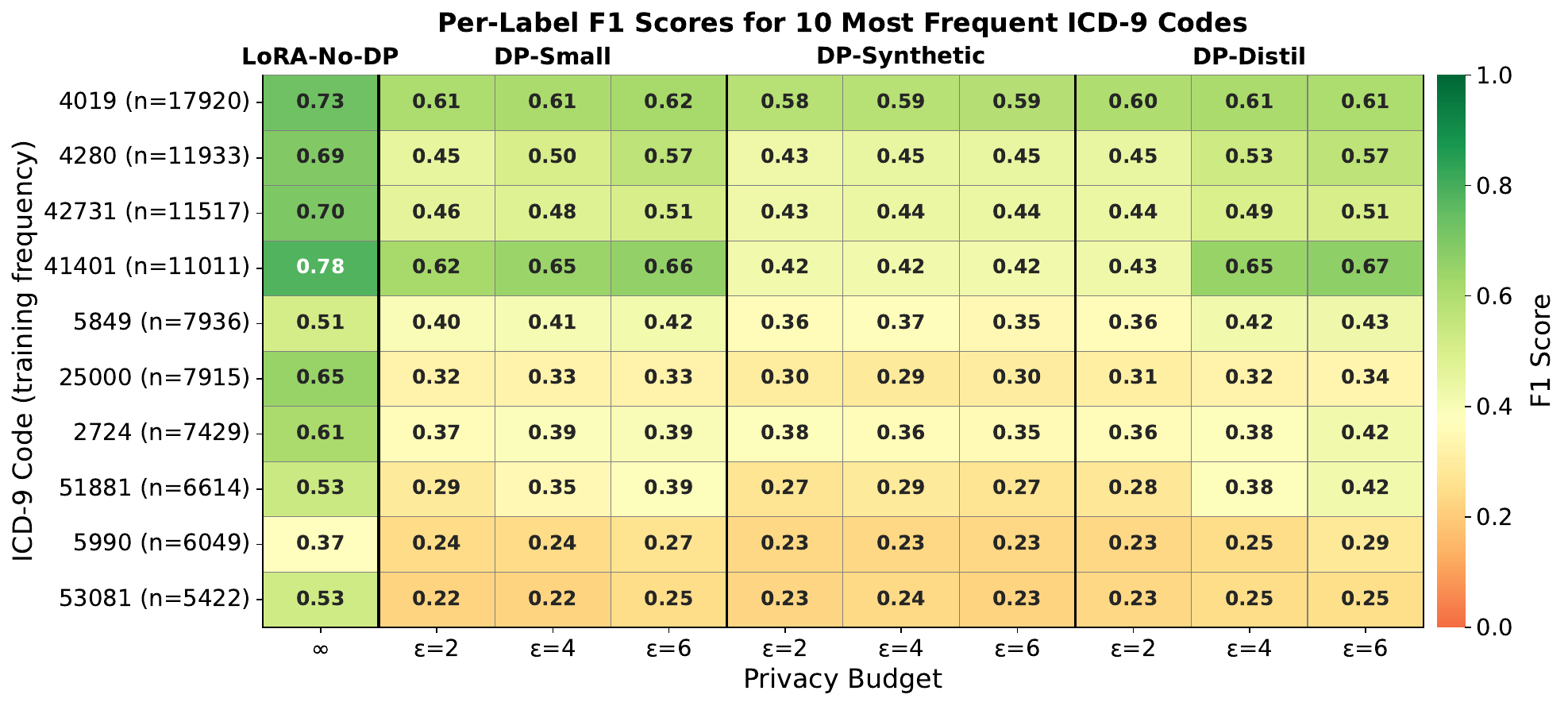}
\caption{Per-label F$_1$ scores for the ten most frequent ICD-9 codes, sorted by decreasing training frequency (n in parentheses). DP-Synthetic consistently underperforms DP-Small and DP-Distil across nearly all codes, while LoRA-No-DP shows irregular memorisation patterns.}
\label{fig:per-label-heatmap}
\end{figure}

Figure~\ref{fig:per-label-heatmap} displays F$_1$ scores for the ten most frequent ICD-9 codes, revealing how privacy mechanisms affect individual diagnoses. All three DP methods show general correlation with code frequency, with performance declining for rarer codes. DP-Small and DP-Distil exhibit similar patterns, with a notable peak at code 41401 (coronary atherosclerosis, fourth most frequent, n=11,011), achieving F$_1$ scores of 0.66 and 0.67 respectively at $\varepsilon=6$.

DP-Synthetic shows consistent underperformance across nearly all codes compared to other DP methods. While it exhibits smooth frequency-dependent degradation—declining from 0.59 (code 401.9) to 0.23 (code 53081) at $\varepsilon=6$—its per-code F$_1$ scores remain systematically lower than DP-Small and DP-Distil regardless of code frequency or privacy budget.

LoRA-No-DP shows the most irregular pattern, with substantial variation unexplained by frequency alone. It achieves 0.78 for code 41401 but only 0.37 for code 5990 (n=6,049).

\subsection{Key observations}

\begin{enumerate}
    \item \textbf{Knowledge distillation outperforms direct DP-SGD ($\varepsilon \geq 4$).} High-capacity teachers retain higher utility under DP-SGD, and their soft labels allow students to learn cleaner decision boundaries than direct DP-SGD on the 1B model.
    \item \textbf{DP-Synthetic fails due to low-fidelity text generation.} Even at relaxed budgets, synthetic data do not capture the clinical signal required for accurate coding.
    \item \textbf{LoRA provides weak implicit privacy.} Low-rank updates slightly constrain memorisation but cannot substitute for formal DP guarantees.
    \item \textbf{Privacy metrics saturate quickly.} Once DP noise is introduced, further relaxation of $\varepsilon$ yields utility gains but negligible empirical privacy degradation.
\end{enumerate}

\section{Discussion}

\subsection{Why knowledge distillation wins}

Among all differentially private methods, DP-Distil achieves the highest utility at moderate and relaxed privacy budgets. Its advantage stems from the ability of high-capacity teacher models to learn more robust representations under DP-SGD despite the injected noise, and to transfer these representations to a smaller student, consistent with prior observations that larger models can maintain higher utility under DP-SGD at fixed privacy budgets \citep{kamath2022dplm}. The use of soft labels plays a central role: rather than exposing the student to one-hot class assignments derived from noisy teacher predictions, the continuous logit distribution encodes relative class likelihoods, offering richer supervision and acting as a regulariser. This allows the student to recover cleaner decision boundaries even when the teachers' parameters are perturbed by privacy noise.

At very tight privacy budgets ($\varepsilon = 2$), DP-Small outperforms DP-Distil; the latter splits its budget between two teachers ($\varepsilon/2$ each), leaving insufficient signal for high-fidelity distillation. As $\varepsilon$ increases, the teachers' larger capacity enables them to tolerate DP noise more effectively. By $\varepsilon = 6$, the 1B student achieves Micro-F$_1$ of 0.33, slightly exceeding the 3B teachers (0.32), consistent with regularisation benefits in prior distillation literature \citep{furlanello2018bornneuralnetworks}. As shown in Table~\ref{tab:teacher-student-comparison}, students consistently match or outperform their teachers across all privacy budgets, with the performance gap widening at higher $\varepsilon$ values where teachers have cleaner signals to transfer.

Taken together, these results highlight that deploying the 3B DP-trained teachers directly would offer no accuracy benefit over the 1B students while nearly doubling memory and latency (Table~\ref{tab:inference-appendix}). Distillation therefore provides both higher utility and a more efficient deployed model, with lower latency and memory requirements, reinforcing its practicality for real-world clinical settings.

\subsection{Why synthetic data fails}

DP-Synthetic consistently underperforms, plateauing around Micro-F$_1$ of 0.22. Since its training procedure mirrors DP-Small but uses DP-generated text, this gap isolates synthetic data fidelity as the limiting factor. The DP generator struggles to reproduce the statistical and lexical diversity of true discharge summaries, leading to label-distribution drift and loss of rare diagnostic patterns.

As shown in Figure~\ref{fig:per-label-heatmap}, this underperformance extends across nearly all individual codes—even the most frequent with ample training examples—confirming that synthetic data lack essential clinical signal rather than merely failing to cover rare patterns.

\subsection{LoRA's modest implicit privacy}

LoRA-No-DP achieves MIA AUC of 0.57, providing measurable but incomplete privacy protection. This represents a 13\% relative increase in attack success over random guessing (AUC = 0.5), partially validating that low-rank constraints inherently limit memorisation \citep{malekmohammadi2025lowrankadaptationsecretlyimitates}. However, the gap to formal DP methods (AUC 0.49--0.51) confirms that LoRA alone cannot replace rigorous privacy guarantees in high-stakes clinical settings.

\subsection{Limitations}

Our study has several limitations. The 512-token truncation captures roughly 17\% of each discharge summary, potentially omitting diagnostic context. Focusing on the 50 most frequent ICD-9 codes may not reflect performance on rare conditions. Our privacy audit evaluates only logit-based membership-inference attacks; other attack classes (e.g., attribute or extraction) remain outside scope. Finally, experiments used a single random seed and limited hyperparameter tuning. Despite these constraints, the observed trends remain robust and internally consistent across metrics and privacy budgets, supporting our central claim that architectural choices, alongside the privacy budget, substantially shape the privacy–utility trade-off.

\subsection{Practical implications}

Taken together, our results provide the first controlled evidence of how privacy mechanisms reshape model performance in clinical NLP at fixed capacity. When strict privacy is required ($\varepsilon \leq 2$), DP-Small remains the simplest and most compute-efficient option. For moderate budgets ($\varepsilon \geq 4$), DP-Distil achieves the best balance of accuracy and protection, recovering up to 63\% of the non-private baseline's performance whilst maintaining near-random MIA outcomes (AUC $\approx$ 0.5). DP-Synthetic is currently unsuitable for deployment, and LoRA should be treated purely as a non-DP upper bound.

Overall, knowledge distillation with differentially private teachers emerges as the most practical path toward deployable, privacy-preserving clinical language models, offering a clear direction for healthcare institutions seeking to combine diagnostic utility with rigorous confidentiality guarantees.

Future work should extend these comparisons to multi-hospital datasets and longer clinical narratives, establishing benchmarks for truly deployable privacy-preserving clinical language models.

\begin{ack}
This work was conducted as part of an MSc thesis at Imperial College London, supervised by Dr Andrew Duncan. We acknowledge Andy Thomas (Head of Research Computing, Imperial Mathematics) for providing access to computing resources, and the MIMIC-III team and PhysioNet for making the clinical database publicly available. Code is available at \url{https://github.com/mathieu-dufour/dp-clinical-coding}.
\end{ack}

\bibliographystyle{plainnat}
\bibliography{refs}

\newpage
\appendix
\section{Supplementary material}

\subsection{Detailed dataset statistics}

MIMIC-III discharge summaries were filtered to 49,426 admissions with at least one top-50 ICD-9 code, split 80/10/10\% into training (44,778 notes from 39,540 admissions), validation (5,624 notes), and test (5,586 notes) sets. Notes average 2,993 tokens (before truncation to 512) and 4.5 codes. Top-5 codes: 401.9 (hypertension, n=17,920), 428.0 (heart failure, n=11,933), 427.31 (atrial fibrillation, n=11,517), 414.01 (coronary atherosclerosis, n=11,011), 584.9 (acute kidney failure, n=7,936).

\subsection{Full training details}

All models trained for maximum 18 epochs with early stopping (patience=3). DP methods used SGD with momentum 0.9 and learning rates of 1.5e-3 for 1B models and 2e-4 to 3e-4 for 3B models. Non-DP methods used AdamW with cosine scheduling and learning rates of 5e-4 to 7.5e-4. Gradient clipping norms: C=1.0 (1B), C=0.7 (3B). Batch sizes were maximised within 48GB GPU memory constraints. Synthetic generation used nucleus sampling (p=0.9, temperature=0.8).

\subsection{Training efficiency and resource requirements}

\begin{table}[h!]
\centering
\caption{Wall-clock training time by pipeline using a single RTX 6000 Ada GPU (48GB VRAM) with maximum batch sizes. Times in hours.}
\label{tab:train-time-appendix}
\small
\begin{tabular}{llrrrrrr}
\toprule
\textbf{Pipeline} & \textbf{$\varepsilon$} & \textbf{Gen.} & \textbf{Synth.} & \textbf{Teacher} & \textbf{Student/} & \textbf{Total} & \textbf{Peak} \\
 & & \textbf{Train} & \textbf{Gen.} & \textbf{Train} & \textbf{Classifier} & \textbf{(h)} & \textbf{VRAM (GB)} \\
\midrule
LoRA-No-DP & $\infty$ & -- & -- & -- & 3.07 & 3.07 & 41.11 \\
\midrule
DP-Small & 2 & -- & -- & -- & 4.91 & 4.91 & 42.24 \\
DP-Small & 4 & -- & -- & -- & 5.60 & 5.60 & 42.24 \\
DP-Small & 6 & -- & -- & -- & 9.56 & 9.56 & 42.24 \\
\midrule
DP-Synthetic & 2 & 8.45 & 3.33 & -- & 2.67 & 14.45 & 41.11 \\
DP-Synthetic & 4 & 8.72 & 3.33 & -- & 3.16 & 15.21 & 41.11 \\
DP-Synthetic & 6 & 8.53 & 3.33 & -- & 2.68 & 14.55 & 41.11 \\
\midrule
DP-Distil & 2 & 10.86 & 3.33 & 18.82 & 5.06 & 38.06 & 41.11 \\
DP-Distil & 4 & 11.05 & 3.33 & 21.59 & 2.34 & 38.31 & 41.11 \\
DP-Distil & 6 & 11.34 & 3.33 & 29.93 & 4.70 & 49.29 & 41.11 \\
\bottomrule
\end{tabular}
\end{table}

\begin{table}[h!]
\centering
\caption{Inference metrics (mean across all 1B classifiers and 3B DP-Distil teachers) measured on a single RTX 6000 Ada GPU with a batch size of 64 and a maximum sequence length of 512 tokens.}
\label{tab:inference-appendix}
\small
\begin{tabular}{lcccc}
\toprule
\textbf{Model} & \textbf{Latency} & \textbf{Throughput} & \textbf{Peak VRAM} & \textbf{Adapter} \\
 & \textbf{(ms/example)} & \textbf{(tokens/s)} & \textbf{(MB)} & \textbf{(MB)} \\
\midrule
1B classifier & 13.06 & 38,376 & 4,432 & 520 \\
3B classifier (DP-Distil teacher) & 35.65 & 14,054 & 8,462 & 1,529 \\
\bottomrule
\end{tabular}
\end{table}

\subsection{Code availability}
Complete code for data preprocessing, model training, and evaluation is available at \url{https://github.com/mathieu-dufour/dp-clinical-coding} under a CC-BY-4.0 licence. The repository includes a clear README describing how to reproduce results, and numbered scripts to run in sequence to perform the full research computations.

\end{document}